\documentclass{article}

\usepackage{PRIMEarxiv}

\usepackage[utf8]{inputenc} 
\usepackage[T1]{fontenc}    
\usepackage{hyperref}       
\usepackage{url}            
\usepackage{booktabs}       
\usepackage{amsfonts}       
\usepackage{nicefrac}       
\usepackage{microtype}      
\usepackage{lipsum}
\usepackage{fancyhdr}       
\usepackage{graphicx}       
\graphicspath{{fig/}}     

\title{ CT-Conditioned Diffusion Prior with Physics-Constrained Sampling for PET Super-Resolution

}

\author{
\textbf{Liutao Yang, Zi Wang, Peiyuan Jing, Xiaowen Wang, Javier A. Montoya-Zegarra,} \\
\textbf{Kuangyu Shi, Daoqiang Zhang, Guang Yang*}}
\begin{document}
%

%


\maketitle              
\begingroup
\renewcommand\thefootnote{}
\footnotetext{
Liutao Yang, Zi Wang, Peiyuan Jing, Xiaowen Wang, and Guang Yang are with the Department of Bioengineering and Imperial-X, Imperial College London, London, U.K. 
Daoqiang Zhang and Liutao Yang are also with the College of Artificial Intelligence, Nanjing University of Aeronautics and Astronautics, Nanjing, China. 
Guang Yang is also with the National Heart and Lung Institute, Imperial College London, London, U.K., and the Cardiovascular Research Centre, Royal Brompton Hospital, London, U.K. 
Peiyuan Jing and Javier A. Montoya-Zegarra are with the School of Engineering, Zurich University of Applied Sciences, Winterthur, Switzerland. 
Kuangyu Shi is with the Department of Nuclear Medicine, Inselspital, University of Bern, Bern, Switzerland. 
Corresponding author: Guang Yang (e-mail: l.yang23@imperial.ac.uk; g.yang@imperial.ac.uk).
}
\endgroup
\begin{abstract}
PET super-resolution is highly under-constrained because paired multi-resolution scans from the same subject are rarely available, and effective resolution is determined by scanner-specific physics (e.g., PSF, detector geometry, and acquisition settings). This limits supervised end-to-end training and makes purely image-domain generative restoration prone to hallucinated structures when anatomical and physical constraints are weak. We formulate PET super-resolution as posterior inference under heterogeneous system configurations and propose a CT-conditioned diffusion framework with physics-constrained sampling. During training, a conditional diffusion prior is learned from high-quality PET/CT pairs using cross-attention for anatomical guidance, without requiring paired LR--HR PET data. During inference, measurement consistency is enforced through a scanner-aware forward model with explicit PSF effects and gradient-based data-consistency refinement. Under both standard and OOD settings, the proposed method consistently improves experimental metrics and lesion-level clinical relevance indicators over strong baselines, while reducing hallucination artifacts and improving structural fidelity.


\end{abstract}
\section{Introduction}
Positron emission tomography (PET) is essential for functional imaging in oncology, neurology, and cardiology, but image quality is fundamentally limited by counting statistics, detector response, and acquisition geometry~\cite{rahmim2008petvsspect,reader2007advancespet,cherry2018totalbody}. Increasing resolution or reducing dose/time amplifies noise and blur, yielding a persistent sensitivity--resolution trade-off~\cite{reader2007advancespet,cherry2018totalbody}.

PET super-resolution is highly ill-posed because degradation is scanner dependent (e.g., PSF and acquisition configuration), not a fixed downsampling rule. Paired multi-resolution scans are therefore scarce, especially across heterogeneous systems, making supervised LR$\rightarrow$HR mapping difficult to generalize. Classical and iterative reconstruction methods enforce physics and Poisson statistics~\cite{reader2007advancespet,qi2006iterative}, while supervised deep models improve appearance but remain sensitive to domain shift and hallucination under mismatch~\cite{ozaltan2024deep,jiang2023pet,zhou2024super}.

Diffusion-based PET reconstruction combines learned priors with data consistency~\cite{ho2020ddpm,song2021score,chung2023dps,singh2024scorepet,webber2024likelihood,bae2023efficient,hashimoto2025ddip}, but three gaps remain for PET super-resolution: limited cross-modal anatomical guidance, weak modeling of heterogeneous scanner operators, and predominantly image-domain restoration with limited measurement domain guarantees~\cite{jiang2023pet,zhou2024super}. Because reconstructed appearance entangles system response and post-processing, robust PET super-resolution should be constrained by the low-quality measurement and its scanner-specific forward model.
\begin{figure}[t]
    \centering
    \includegraphics[width=0.95\linewidth]{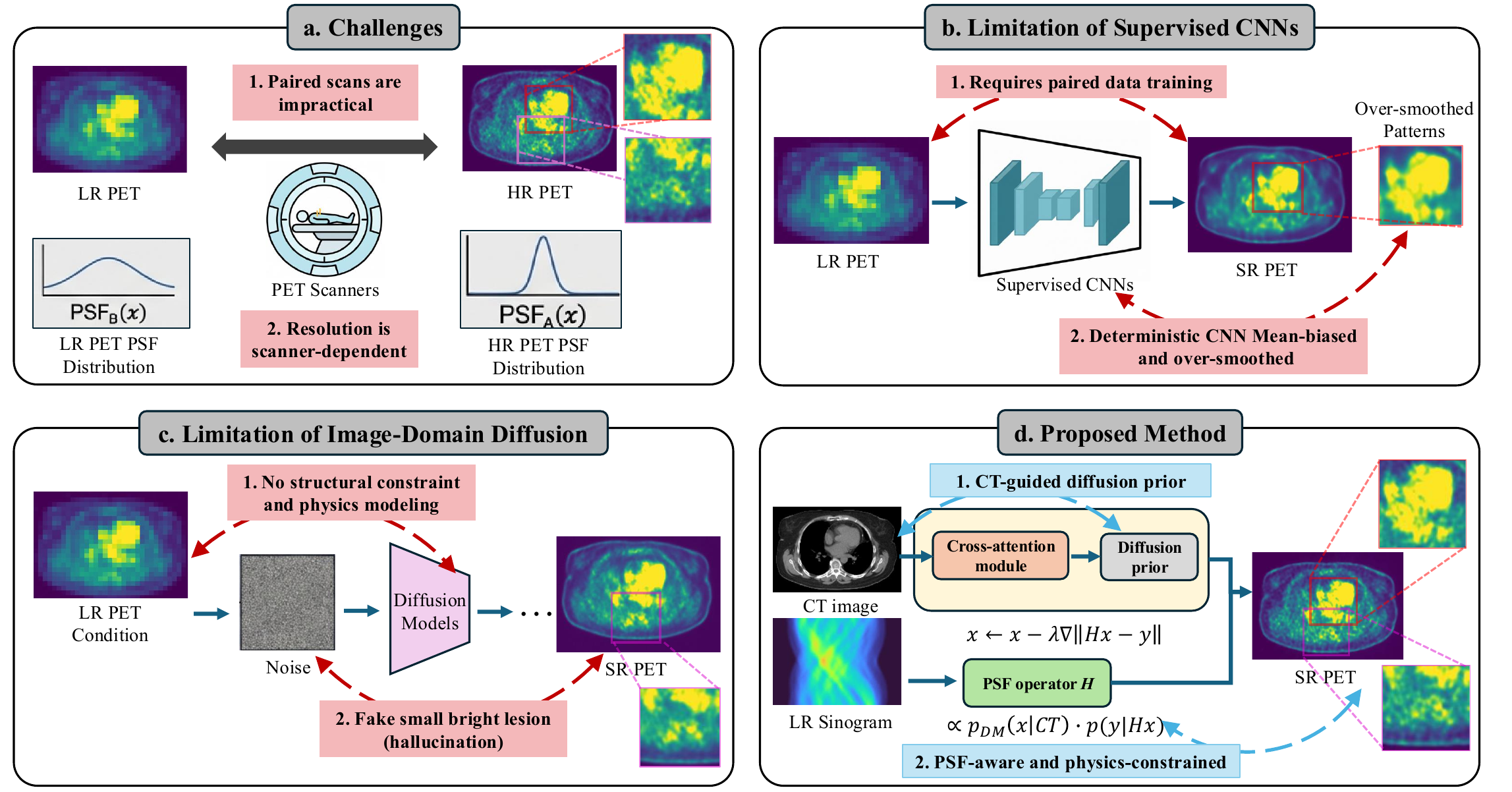}
    \caption{Conceptual motivation and method positioning for PET super-resolution.
(a) PET super-resolution is ill-posed due to scanner-dependent degradation and limited paired data.
(b) Supervised CNN mapping depends on paired LR--HR data and tends to over-smooth details.
(c) Image-domain diffusion can recover texture but may introduce hallucinations without structural/physics constraints.
(d) Our method combines CT-conditioned priors with scanner-aware, PSF-aware consistency for faithful and physically consistent reconstruction.
}
    \label{fig:motivation}
\end{figure}
To address these limitations, we reformulate PET super-resolution as posterior inference under heterogeneous scanner configurations. The proposed framework combines a CT-conditioned diffusion prior with physics-constrained sampling: CT cross-attention provides anatomical guidance to suppress hallucination, while PSF-aware measurement-domain data consistency constrains samples to scanner-consistent solutions. This design targets the key failure modes of prior methods by jointly improving structural fidelity, physical plausibility, and robustness across standard and out-of-distribution degradation settings.

The main contributions of this work are summarized as follows:
\begin{itemize}
    \item We formulate PET super-resolution as a posterior inference problem that progressively constrains the solution space using a CT-conditional diffusion prior and PSF-based data consistency.
    \item We highlight the distinction between underlying activity estimation and reconstructed image appearance, enabling physically consistent reconstruction without paired multi-resolution datasets.
    \item We demonstrate that the proposed framework outperforms supervised end-to-end models and image-domain diffusion approaches in both structural fidelity and quantitative accuracy.
\end{itemize}

\section{Methods}
\begin{figure}[t]
\centering
\includegraphics[width=0.95\columnwidth]{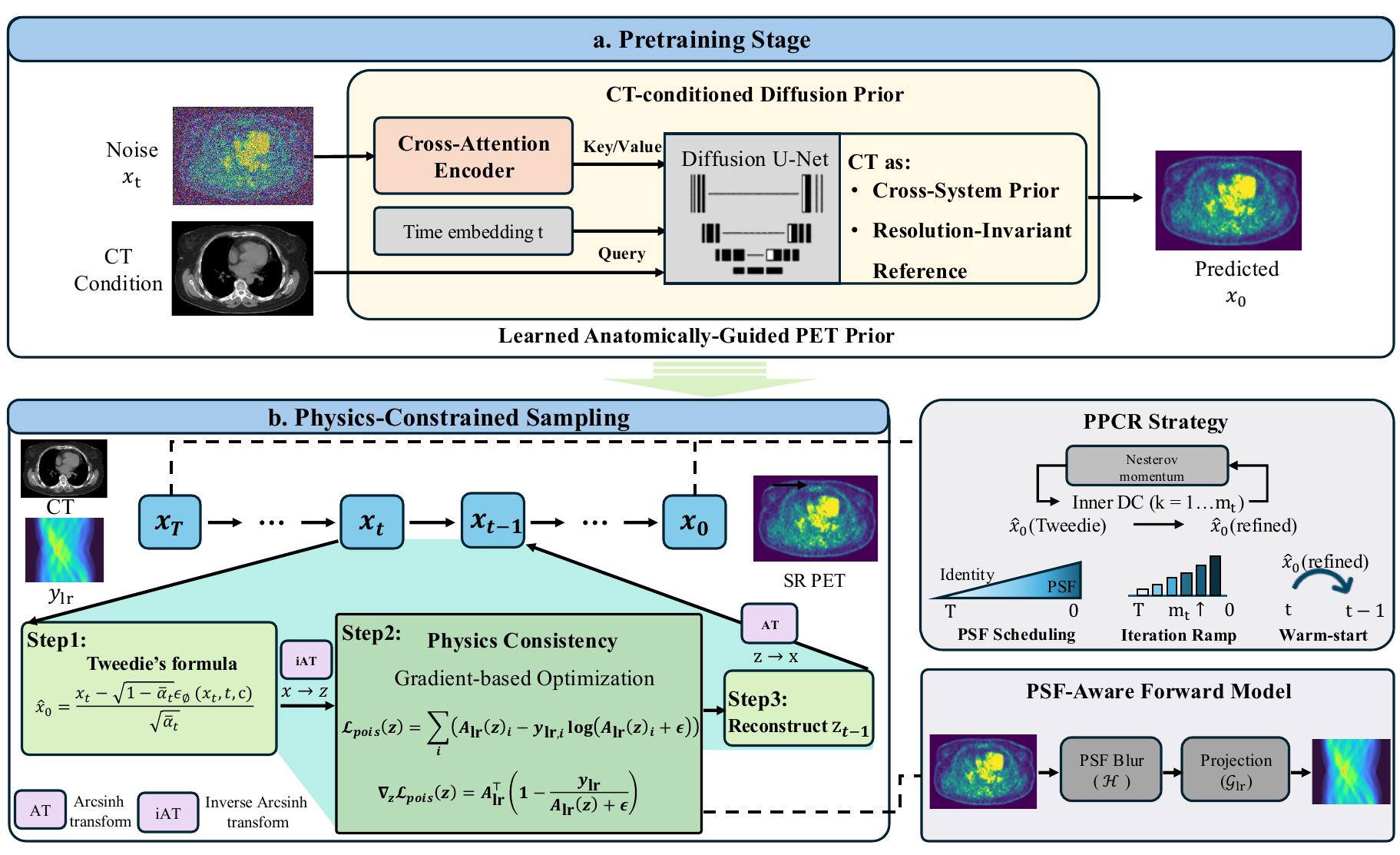}
\caption{
Overview of the proposed CT-guided diffusion framework.	\textbf{Pretraining (a):} learn a CT-conditioned diffusion prior from high-quality PET/CT.\textbf{Inference (b):} start from noisy PET, iteratively denoise, and apply measurement-domain data-consistency updates. \textbf{PPCR:} progressively enforces physics constraints (late activation and coarse-to-fine PSF) to produce a scanner-consistent high-resolution PET output.
}
\label{fig:framework}
\end{figure}
\subsection{Problem Formulation}

Let $z \in \mathbb{R}^N$ denote the tracer activity distribution
in physical units.
For scanner configuration $k$, PET measurements follow:

\begin{equation}
y_k \sim \mathrm{Poisson}(A_k z + b_k),
\end{equation}
where $A_k$ denotes the scanner-specific forward operator and $b_k$ represents background contributions (e.g., random and scatter events).
In the following, the low-quality acquisition 
is denoted by $y_{\mathrm{lr}}$ and its forward operator by $A_{\mathrm{lr}}$.

Rather than learning an explicit mapping between resolutions,
we formulate PET super-resolution as posterior inference:

\begin{equation}
p(z \mid y_{\mathrm{lr}}, c)
\propto
p_\phi(z \mid c)\, p(y_{\mathrm{lr}} \mid z),
\end{equation}

where $p_\phi(z \mid c)$ is a learned structural prior,
with $c$ denoting the co-registered CT anatomy used for conditioning; and $p(y_{\mathrm{lr}} \mid z)$ is defined by scanner-specific physics (including PSF effects, projection geometry, and Poisson noise statistics).

\subsection{Score-Based CT-Conditional Prior}

We model $p_\phi(z \mid c)$ using a conditional diffusion model
\cite{ho2020ddpm,song2021score}.
Diffusion models learn the score function
$\nabla_x \log p_\phi(x \mid c)$
of the data distribution in a normalized image space.

Given a physical activity map $z$,
we map it to model space via an arcsinh transform:

\begin{equation}
x = \frac{\mathrm{asinh}(z/s)}{\kappa}.
\end{equation}

Here, $\kappa$ is a scaling constant. The arcsinh transform stabilizes PET's high dynamic range: it behaves approximately logarithmically at high uptake while remaining well behaved near zero.

During training, Gaussian noise is added to $x$,
and a neural network $\epsilon_\phi(x_t,t,c)$
is trained to predict the injected noise.
This is equivalent to learning the conditional score
under the diffusion framework \cite{song2021score}.

CT conditioning is incorporated via cross-attention in a Transformer-style formulation~\cite{vaswani2017attention},
providing anatomical guidance.
The prior is learned exclusively from high-quality PET/CT pairs,
without requiring degradation modeling.

\subsection{Physics-Constrained Likelihood Modeling}

The measurement likelihood is defined via the forward operator:

\begin{equation}
A_{\mathrm{lr}}(z) = \mathcal{G}_{\mathrm{lr}}(\mathcal{H} z) + b_{\mathrm{lr}},
\end{equation}

where:
 $z \in \mathbb{R}^N$ is the activity map (MBq/mL), $\mathcal{H}$ is a PSF operator (Gaussian, scanner-specific FWHM), $\mathcal{G}_{\mathrm{lr}}$ is the low-quality scanner measurement operator, and $b_{\mathrm{lr}}$ models background (random/scatter), yielding a unified scanner-aware likelihood.

Under a Poisson measurement model,
the negative log-likelihood (up to constants) is:

\begin{equation}
\mathcal{L}_{\mathrm{Pois}}(z)
=
\sum_i \left(A_{\mathrm{lr}}(z)_i - y_{\mathrm{lr},i}\log\left(A_{\mathrm{lr}}(z)_i + \epsilon\right)\right),
\end{equation}

where $\epsilon>0$ is a small numerical constant.
Its gradient is computed analytically as:
\begin{equation}
\nabla_z \mathcal{L}_{\mathrm{Pois}}(z)
=
A_{\mathrm{lr}}^\top\!\left(1 - \frac{y_{\mathrm{lr}}}{A_{\mathrm{lr}}(z)+\epsilon}\right).
\end{equation}

In implementation, we compute $A_{\mathrm{lr}}(z)$ by forward projection and apply adjoint backprojection via $A_{\mathrm{lr}}^\top$; no autograd-through-projector is required.

\subsection{Tweedie-Guided Posterior Refinement with Progressive Physics-Constrained Refinement}

Reverse diffusion sampling can be interpreted
as approximate posterior sampling under a learned score prior
\cite{song2021score,chung2023dps}.

At timestep $t$, the diffusion model predicts noise,
from which a denoised estimate can be obtained
via Tweedie’s formula \cite{efron2011tweedie}:

\begin{equation}
\hat{x}_0 =
\frac{x_t - \sqrt{1-\bar{\alpha}_t}
\,\epsilon_\phi(x_t,t,c)}
{\sqrt{\bar{\alpha}_t}}.
\end{equation}

The quantity $\hat{x}_0$ serves as a Tweedie’s estimate
of the clean image under the learned prior.

To incorporate measurement information,
we refine this Tweedie estimate using likelihood-gradient descent in activity space:

\begin{equation}
z^{\mathrm{refined}}
=
z^{\mathrm{Tweedie}}
-
\eta_t
\nabla_z \mathcal{L}_{\mathrm{Pois}}(z^{\mathrm{Tweedie}}).
\end{equation}

To improve convergence under PSF-based DC, we use Progressive Physics-Constrained Refinement (PPCR): coarse-to-fine PSF scheduling (no-PSF early, full-PSF late), increasing inner DC iterations over timesteps, and Nesterov acceleration with warm-start~\cite{friedman2010glmnet,boyd2011admm}.

Formally, let $m_t$ denote the number of inner DC iterations
and let $\mathcal{H}_t$ denote the scheduled PSF operator
(identity in early steps, full PSF in late steps).
The DC update at timestep $t$ is:

\begin{equation}
z_{t}^{(k+1)}
=
z_{t}^{(k)}
-
\eta_t\,\nabla_z \mathcal{L}_{\mathrm{Pois}}^{\mathcal{H}_t}(z_{t}^{(k)}),
\quad k=1,\dots,m_t,
\end{equation}

with $m_t$ increasing over $t$ (e.g., linear ramp), allocating more DC computation near the final high-fidelity stage. Early no-PSF updates improve conditioning, and late full-PSF updates restore measurement-faithful fine detail. This alternating denoising--DC scheme yields a stable and physically consistent posterior approximation for CT-guided PET super-resolution~\cite{song2021score,chung2023dps,singh2024scorepet}.
\section{Experiments}

\subsection{Experimental Setup}

\subsubsection{Dataset} Experiments were conducted on the FDG-PET-CT-Lesions dataset \cite{gatidis2022whole} (TCIA).
For this study, 120 cases were randomly split into 100/10/10 for training/validation/testing.
	 For supervised comparison methods, paired LR--HR training data are generated using simulated LR data from the training split. Our method is trained without simulated LR--HR PET pairs. All methods are then evaluated on the same unified simulated test settings for fair comparison. Detailed simulation settings are described in Degradation Protocol.

\subsubsection{Degradation Protocol} To evaluate super-resolution performance under controlled physical conditions, low-resolution measurements were simulated using a physics-based degradation pipeline.
The forward process incorporates Gaussian PSF blurring, projection geometry, angular and radial rebinning, and dose reduction.
Two degradation configurations were designed to assess both standard and out-of-distribution (OOD) scenarios. 
The corresponding parameters are summarized in Table~\ref{tab:degradation}.
The standard setting corresponds to approximately $\times4$ resolution enhancement relative to the original PET spacing.
The OOD setting introduces stronger PSF blurring and ultra-low dose conditions, resulting in approximately $\times6$ super-resolution difficulty.
\begin{table}[t]
\centering
\caption{Degradation settings for standard and out-of-distribution (OOD) evaluation.}
\label{tab:degradation}
\begin{tabular}{lcccccc}
\toprule
Setting & FWHM & Dose & Angular & Radial & Target Spacing & SR Factor \\
\midrule
Standard & 8 mm & 10\% & $\times$2 & $\times$2 & 8 mm & $\times$4 \\
OOD & 12 mm & 5\% & $\times$3 & $\times$2 & 12 mm & $\times$6 \\
\bottomrule
\end{tabular}
\end{table}

\subsubsection{Implementation Details} The proposed model was trained in a 2D slice-wise manner.
A CT-conditioned cross-attention diffusion UNet was pretrained to learn the anatomically guided PET prior.
During inference, DDIM sampling used 50 steps with PPCR: no-PSF DC in steps 1--35, full-PSF DC in steps 36--50, inner DC iterations linearly ramped from 2 to 20, and Nesterov ($\mu=0.9$) with warm-start ($\alpha=0.3$) at DC step size 0.05.

\subsection{Comparison with State-of-the-Art}

\subsubsection{Compared Methods} The proposed method was compared against representative supervised and unsupervised PET super-resolution approaches. \textbf{Supervised methods} include 
UNet~\cite{ronneberger2015unet},
SRCNN~\cite{dong2014srcnn},
and DeepSR~\cite{ozaltan2024deep}.
    \textbf{Unsupervised methods} include 
ScorePET~\cite{singh2023score},
PETDM~\cite{jiang2023pet},
and SRDPM~\cite{zhou2024super}.
All supervised models were retrained on the same training split.
Unsupervised and diffusion-based methods were implemented following the configurations described in their respective publications.

\subsubsection{Quantitative and Qualitative Evaluation} 
\begin{figure}[t]
\centering
\includegraphics[width=1\columnwidth]{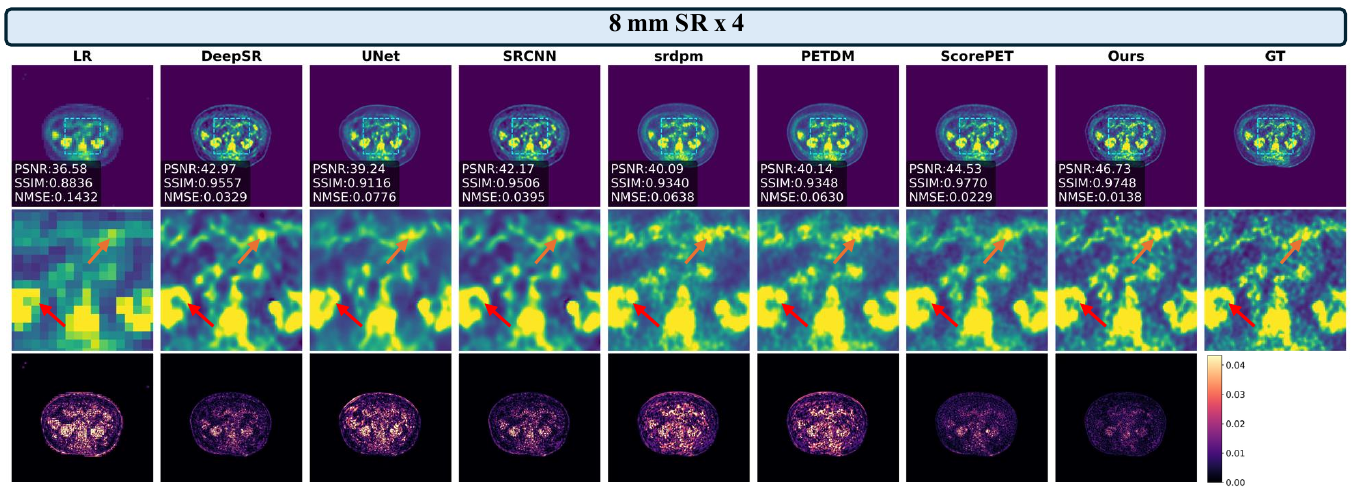}
\caption{Qualitative comparison under the standard setting (8\,mm, SR$\times$4).}
\label{fig:qualitative}
\end{figure}

Reconstruction performance was evaluated using PSNR, SSIM, and NMSE. Quantitative results for both standard and OOD settings are summarized in Table~\ref{tab:compare_all}. Under standard (8\,mm, SR$\times$4), ScorePET is the strongest baseline, while under OOD (12\,mm, SR$\times$6) all baselines degrade, especially supervised methods. Our method remains best in both settings, indicating stronger fidelity and robustness under increased degradation.
Qualitative examples are shown only for the standard setting in Fig.~\ref{fig:qualitative}, while OOD comparisons are reported in Table~\ref{tab:compare_all}. CNN-based methods over-smooth lesion boundaries, and diffusion baselines recover texture but show less structural stability; in contrast, our method preserves finer details with better quantitative consistency.
\begin{table*}[t]
\centering
\caption{Quantitative comparison under standard (8\,mm, SR$\times$4) and OOD (12\,mm, SR$\times$6) settings. Best results are \textbf{bolded}; second-best are \underline{underlined}.}
\label{tab:compare_all}
\begin{tabular}{lccc|ccc}
\toprule
& \multicolumn{3}{c|}{\textbf{8\,mm (SR$\times$4)}} & \multicolumn{3}{c}{\textbf{12\,mm (SR$\times$6)}} \\
\cmidrule(r){2-4} \cmidrule(l){5-7}
Method & PSNR$\uparrow$ & SSIM$\uparrow$ & NMSE$\downarrow$ & PSNR$\uparrow$ & SSIM$\uparrow$ & NMSE$\downarrow$ \\
\midrule
LR & 36.63$\pm$4.56$^{*}$ & 0.887$\pm$0.04$^{*}$ & 0.09$\pm$0.05$^{*}$ & 34.98$\pm$3.75$^{*}$ & 0.849$\pm$0.05$^{*}$ & 0.09$\pm$0.05$^{*}$ \\
\midrule
DeepSR~\cite{ozaltan2024deep} & 42.49$\pm$3.70$^{*}$ & 0.957$\pm$0.02$^{*}$ & 0.03$\pm$0.03$^{*}$ & 37.27$\pm$3.31$^{*}$ & 0.896$\pm$0.04$^{*}$ & 0.06$\pm$0.05$^{*}$ \\
UNet~\cite{ronneberger2015unet} & 39.50$\pm$4.55$^{*}$ & 0.929$\pm$0.05$^{*}$ & 0.06$\pm$0.07$^{*}$ & 36.35$\pm$3.81$^{*}$ & 0.877$\pm$0.07$^{*}$ & 0.09$\pm$0.10$^{*}$ \\
SRCNN~\cite{dong2014srcnn} & 42.75$\pm$3.70$^{*}$ & 0.960$\pm$0.02$^{*}$ & 0.03$\pm$0.03$^{*}$ & 37.33$\pm$3.32$^{*}$ & 0.899$\pm$0.04$^{*}$ & 0.06$\pm$0.05$^{*}$ \\
SRDPM~\cite{zhou2024super} & 40.30$\pm$3.29$^{*}$ & 0.923$\pm$0.03$^{*}$ & 0.04$\pm$0.05$^{*}$ & 36.74$\pm$3.08$^{*}$ & 0.877$\pm$0.05$^{*}$ & 0.08$\pm$0.05$^{*}$ \\
PETDM~\cite{jiang2023pet} & 41.32$\pm$3.27$^{*}$ & 0.924$\pm$0.03$^{*}$ & 0.04$\pm$0.05$^{*}$ & 36.84$\pm$3.13$^{*}$ & 0.879$\pm$0.05$^{*}$ & 0.08$\pm$0.05$^{*}$ \\
ScorePET~\cite{singh2023score} & \underline{43.97$\pm$3.84}$^{*}$ & \underline{0.970$\pm$0.02}$^{*}$ & \underline{0.02$\pm$0.02}$^{*}$ & \underline{37.85$\pm$3.35}$^{*}$ & \underline{0.912$\pm$0.04}$^{*}$ & \underline{0.05$\pm$0.04}$^{*}$ \\
\midrule
\textbf{Ours} & \textbf{46.84$\pm$3.08} & \textbf{0.980$\pm$0.01} & \textbf{0.01$\pm$0.01} & \textbf{40.66$\pm$2.86} & \textbf{0.945$\pm$0.03} & \textbf{0.03$\pm$0.03} \\
\bottomrule
\end{tabular}
\end{table*}

\begin{table}[t]
\centering
\caption{Ablation study under the standard degradation setting (8\,mm, SR$\times$4). 
``CT cond.'' indicates the conditioning strategy. Best results are \textbf{bolded}.}
\label{tab:ablation}
\begin{tabular}{lccccccc}
\toprule
& \multicolumn{4}{c}{Components} & \multicolumn{3}{c}{Metrics} \\
\cmidrule(r){2-5} \cmidrule(l){6-8}
Variant 
& CT cond. 
& DC 
& PSF 
& Sched. 
& PSNR$\uparrow$ 
& SSIM$\uparrow$ 
& NMSE$\downarrow$ \\
\midrule

w/o DC 
& CrossAttn 
&  
&  
&  
& 43.92$\pm$3.45 
& 0.962$\pm$0.02 
& 0.018$\pm$0.02 \\

w/o PSF 
& CrossAttn 
& \checkmark 
&  
& \checkmark 
& 45.11$\pm$3.26 
& 0.971$\pm$0.02 
& 0.014$\pm$0.01 \\

w/o PPCR 
& CrossAttn 
& \checkmark 
& \checkmark 
&  
& 45.78$\pm$3.18 
& 0.975$\pm$0.01 
& 0.012$\pm$0.01 \\

CT Concat 
& Concat 
& \checkmark 
& \checkmark 
& \checkmark 
& 45.94$\pm$3.15 
& 0.976$\pm$0.01 
& 0.011$\pm$0.01 \\

\midrule

\textbf{Ours} 
& CrossAttn 
& \checkmark 
& \checkmark 
& \checkmark 
& \textbf{46.84$\pm$3.08} 
& \textbf{0.980$\pm$0.01} 
& \textbf{0.010$\pm$0.01} \\

\bottomrule
\end{tabular}
\end{table}

\subsubsection{Clinical Relevance Analysis}
\begin{figure}[t]
    \centering
    \includegraphics[width=0.95\columnwidth]{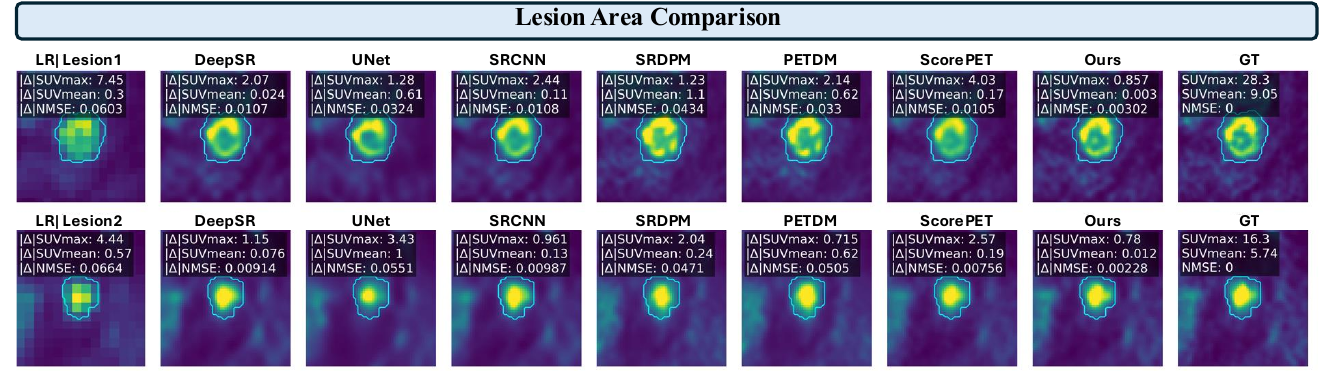}
    \caption{Lesion-level quantitative comparison on two lung cancer cases.
Each lesion reports $\Delta$SUVmax, $\Delta$SUVmean, and lesion NMSE within the lesion mask; lower values indicate better consistency.}
    \label{fig:lesion_quantitative}
\end{figure}
To assess clinical relevance beyond whole-image metrics,
we further performed lesion-level analysis (Fig.~\ref{fig:lesion_quantitative}) on two lung cancer cases.
Across both lesions, our method shows the smallest SUVmax/SUVmean deviations and the lowest lesion NMSE among compared methods, indicating improved lesion-specific quantitative fidelity beyond visual enhancement.

\subsection{Ablation Study}

We conducted ablation experiments under the standard setting (8\,mm, SR$\times$4) to evaluate the contributions of DC, PSF modeling, PPCR, and CT conditioning (Table~\ref{tab:ablation}). Evaluated variants include \textbf{w/o DC} (remove physics-based data consistency), \textbf{w/o PSF} (keep DC but remove explicit PSF modeling), \textbf{w/o PPCR} (keep DC/PSF but disable progressive scheduling), and \textbf{CT Concat} (replace cross-attention with feature concatenation).

Removing DC causes the largest drop, showing that measurement consistency is the dominant factor. Removing PSF or PPCR further degrades performance, confirming the importance of physically accurate blur modeling and progressive enforcement. Replacing cross-attention with concatenation also reduces performance, indicating that cross-attention provides stronger CT-guided structural conditioning.

\section{Conclusion}
We proposed a CT-conditioned diffusion framework with physics-constrained sampling for PET super-resolution.
The method combines a cross-attention PET prior with PSF-aware measurement consistency and shows consistent gains over supervised and diffusion baselines under both standard and OOD degradations.
Ablations confirm the roles of data consistency, PSF modeling, and cross-attention conditioning.
Current limitations include 2D slice-wise inference on simulated degradations; future work will extend to fully 3D and dynamic PET with validation on real raw-data pipelines.


\bibliographystyle{unsrt}  
\bibliography{ref}

\end{document}